# LesionAid: Vision Transformers-based Skin Lesion Generation and Classification

Ghanta Sai Krishna[$,*]
Ghanta20102@iiitnr.edu.in

Kundrapu Supriya[Ψ]
Kundrapy20100@iiitnr.edu.in

Mallikharjuna Rao K[†,*]
Assistant professor
rao.mkrao@gmail.com, mallikharjuna@iiitnr.edu.in

Meetiksha Sorgile[#]
Meetiksha20102@iiitnr.edu.in

[$, †, #]Data Science and Artificial Intelligence, [Ψ]Computer Science and Engineering,
International Institute of Information Technology Naya Raipur, India

**Abstract —** Skin cancer is one of the most prevalent forms of human cancer. It is recognized mainly visually, beginning with clinical screening and continuing with the dermoscopic examination, histological assessment, and specimen collection. Deep convolutional neural networks (CNNs) perform highly segregated and potentially universal tasks against a classified fine-grained object. This research proposes a novel multi-class prediction framework that classifies skin lesions based on ViT and ViTGAN. Vision transformers-based GANs (Generative Adversarial Networks) are utilized to tackle the class imbalance. The framework consists of four main phases: ViTGANs, Image processing, and explainable AI. Phase 1 consists of generating synthetic images to balance all the classes in the dataset. Phase 2 consists of applying different data augmentation techniques and morphological operations to increase the size of the data. Phases 3 & 4 involve developing a ViT model for edge computing systems that can identify patterns and categorize skin lesions from the user's skin visible in the image. In phase 3, after classifying the lesions into the desired class with ViT, we will use explainable AI (XAI) that leads to more explainable results (using activation maps, etc.) while ensuring high predictive accuracy. Real-time images of skin diseases can capture by a doctor or a patient using the camera of a mobile application to perform an early examination and determine the cause of the skin lesion. The whole framework is compared with the existing frameworks for skin lesion detection.

## 1. Introduction

A skin lesion is a region of the skin that differs from the surrounding skin in terms of growth pattern or appearance. Skin lesions can be divided into two categories: primary and secondary. Primary skin lesions are abnormal skin disorders that can develop during a person's lifetime or be present at birth. Primary skin lesions that have been handled or inflamed will lead to secondary skin lesions. Some primary skin lesions are macules, papules, nodules, and tumors, whereas some secondary skin lesions are scales, crusts, excoriations, erosions, and ulcers. Cancerous malignant tumors can occasionally be fatal. Melanoma is the most severe type of skin cancer in lesions [1]. The conclusions regarding skin problems are based mostly on dermatologists or dermatologists: their subjective analysis, which is not always guaranteed error-free, or their years of expertise. So, to distinguish between diseases, a very high degree of knowledge and expertise is required. Owing to this massive reliance on professionals and considering their exorbitant costs for physicals and consultations, many people frequently disregard or cannot afford to pay for medical treatment if they have one. Despite worrying figures, skin problems have received very little attention globally [2]. Computer-aided software may identify skin issues, and the outcomes could be more reliable and predictable.

Deep learning has made it possible to solve complicated learning issues that traditional, rule-based methods have needed help with. Deep learning-based algorithms have come close to matching human performance on various challenging computer vision and image classification tasks. As a result, deep learning techniques are often applied in medical imaging for various purposes [3], such as illness detection. However, deep architectures need to learn many training instances to acquire usable representations. Unfortunately, developing large-scale medical imaging datasets (for supervised learning) is more complex than other applications. Due to the need for specific equipment and skilled medical professionals, acquisition and labeling are expensive and time-consuming operations. Data is one of the major drawbacks of modern deep learning and computer vision frameworks. There are countless skin lesions and variations among individuals; hence there is always a lack of skin-related data.

Convolutional Neural Networks (CNN) is regarded as the most widely used deep learning technology in computer vision. It is a neural network with hierarchically stacked layers, with the output of one layer feeding into the input of the next layer. To estimate generative models in an adversarial manner, Goodfellow et al. introduced a unique GAN framework in 2014.[4] There are two primary parts of GAN: the generator and the discriminator. Both are generally neural networks based on generative and discriminative models, respectively. The samples produced by the generator are regarded as negative training examples for the discriminator as it learns to produce realistic samples. On the other hand, the discriminator learns to differentiate between the generator's bogus samples and true ones and, as a result, penalizes the generator for creating irrational samples. GAN stands out for its capacity to produce realism-based synthetic pictures based on the original dataset.

Our work's primary objectives include the following:

- To perform basic image preprocessing techniques such as morphological operations, histogram equalization, etc., to enhance the feature extraction / Region of Interest (ROI).
- Generating synthetic data by training ViTGAN to minimize class imbalance.
- Training a Vision Transformers (ViT) architecture with both synthetic and original data with proper data augmentation techniques to achieve better performance.
- Performing a comparative analysis of the proposed framework with the existing frameworks regarding computational efficiency, performance, and explainable AI (XAI).
- To develop a robust, user-friendly web application that can assist healthcare experts in obtaining a preliminary analysis by detecting and classifying skin lesions on real-time data.

## 1.1. Challenges

- **Existence of Class Imbalance:** The dataset used in this work consists of a class imbalance problem, i.e., the NV class has 6705 lesion images, and MEL class has only 142 lesion images. Because of this, there may be difficulty in generating generalized outputs.
- **Wide range of diseases:** More number of target classes.
- **Compatibility for deployment:** The model must be computationally efficient to be deployed in an edge device.
- **Overfitting:** Due to multiple target classes, the existing works face problems such as overfitting, degrading the entire architecture's performance.

## 2. RELATED WORKS

To automate the process of skin issue diagnosis, researchers are using computer-aided frameworks. The fundamental operations of image capturing, preprocessing, segmentation, feature extraction, and image classification have been accomplished using techniques. They have tried to address difficulties caused by the scarcity of data for deep learning model training. Quan Gan et al., [5] employed picture color and texture features to identify skin diseases. First, the images were preprocessed using median filtering. In order to obtain the image segments, denoise images are rotated. After extracting text features with the GLCM tool, SVM was used to classify the skin conditions herpes, dermatitis, and psoriasis. A methodology for automatic eczema identification and intensity assessment utilizing image processing and computer algorithms was proposed by Md. Nafiul Alam et al., [6]. By enabling users to upload a picture of the infected body surface, the framework could identify and gauge the intensity of eczema. Our framework used picture segmentation, feature extraction, and statistical classification to distinguish between moderate and severe eczema. An intensity index was given to the photograph when the type of eczema was determined. Deep learning algorithms were later employed in studies to categorize skin conditions. AlexNET, a pre-trained CNN model to extract the features, was devised, implemented, and evaluated by Nazia Hameed et al., [8] to categorize skin lesion images into one of five classes: healthy, acne, eczema, benign, or malignant melanoma. The SVM classifier is employed to perform the classification, and an overall accuracy of 86.21% was attained. Parvathaneni Naga Srinivasu et al., [9] used MobileNet V2 and Long Short-Term Memory to categorize skin illnesses, both of which rely on deep learning. The rate of illness growth was calculated using a co-occurrence matrix at different gray levels. The HAM10000 skin disease dataset has shown the technique to be 85% accurate. The major drawback of this work is that it causes overfitting problems.

Nowadays, researchers are becoming interested in a different technology called generative adversarial network (GAN) since it can simulate complex real-world image data. Additionally, it can make skewed datasets more balanced. However, only some applications have implemented GAN for binary classification and data augmentation. According to Qin et al., [6], creating artificial samples for vascular lesions (142 input photos) results in a realistic-looking lesion location. In other samples, the skin's surface texture around the lesion is still fuzzy and devoid of contrast. Melanoma photos have the same effect. The most representative photos of melanocytic nevus (6705 images) have concentrated color, clean edges, and regular shapes. Therefore, producing more training samples will result in higher-quality synthetic samples. To create skin problem data, Maleika et al. , [10], created a deep generative adversarial network (DGAN) multi-class classifier that can extract the genuine representation of the data from the provided images. We have created a multi-class solution as opposed to the conventional two-class classifier, and to address the class-imbalanced dataset, we have taken photographs from various web datasets.

The improvement of the DCGAN model's stability during the training phase has been one of our development's primary challenges. By enhancing the training datasets using the conventional rotation, etc. techniques, we created two CNN models in parallel based on the architecture of ResNet50 and VGG16 to assess the accuracy of GAN. In SkinAid [12], the authors suggested an application that can help the dermatologist receive an initial assessment of identifying, categorizing, and keeping track of the skin lesion while also educating the user about skin care. Several skin lesions are classified by SkinAid, which also provides information on them. They have investigated Generative Adversarial Networks (GANs) to improve and expand the dataset.

Additionally, they developed a Deep Convolutional Neural Network (CNN) model for edge cloud applications that can instantly identify and categorize skin lesions from the patient's skin image. The total accuracy of the model, which detects and categorizes seven distinct skin lesions, is 92.2%. A healthcare professional can use a mobile application camera to take real-time pictures of skin lesions and upload them to the Internet of Medical Things (IoMT) platform for remote monitoring.

## 3. Research methodology

LesionAid provides the patient with a mobile app that can detect and classify skin lesions by exploiting smartphones' computation power and digital lens. This forms the next wave of advancement in intelligent skincare. The proposed methodology is categorized into 4 phases; conventional data Augmentation and GANs, Predictive modeling with Vision Transformers, Experimental Analysis and Explainable AI (XAI), and Web Application Deployment. The fig 1 describes the functional model of the proposed solution. The below fig 2 depicts the end-to-end architecture that will be employed in this work and subsequently, it is explained in detail in the following subsections.

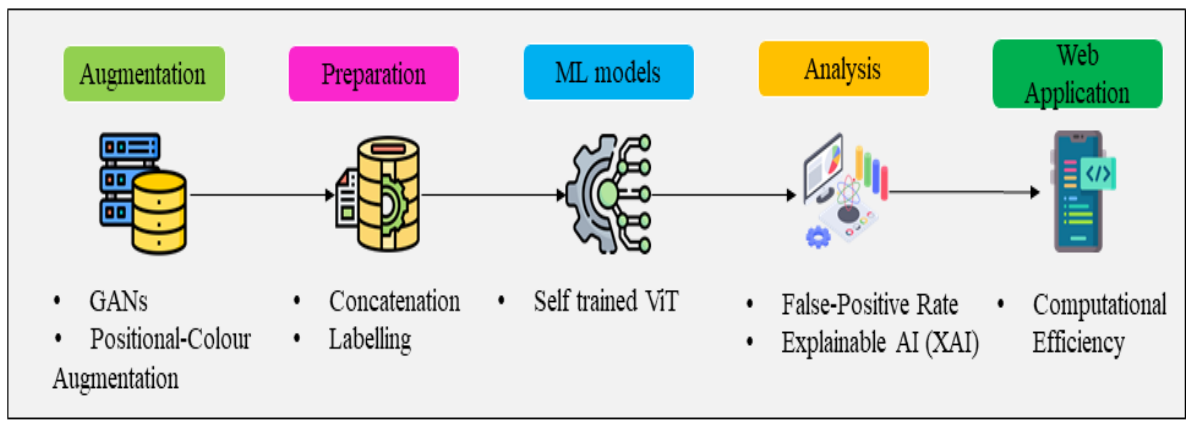

Fig 1. The Functional Model of the proposed solution

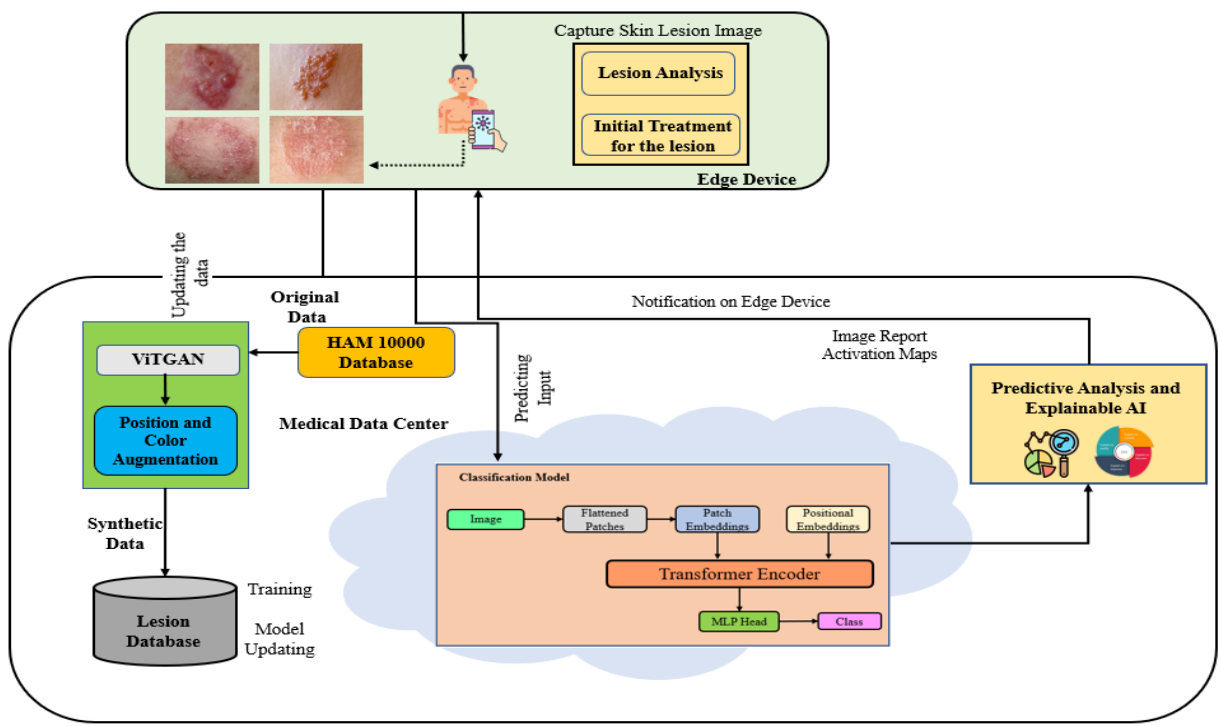

Fig 2. Conceptual Diagram of the Proposed Solution

### 3.1. Dataset Description

This current work is evaluated on the HAM10000 ("Human Against Machine 10000") dataset, a collection of diverse skin diseases and dermoscopic images. There are 10,015 skin lesion samples in this dataset, including seven kinds of lesions open to the general public. The fig 3 depicts the existence of class imbalance among all the classes present in the dataset. The significant drawbacks of this dataset are: a small class size, is imbalanced, and is directly trained on vision transformer architectures since it would over-fit and fail to produce generalized results, further lowering the accuracy. The training set (HAM 10000) is used to train ViTGAN and generate synthetic data (lesion images) to overcome this problem. With the help of the generated data, we can overcome the problem of imbalanced classes and limited data difficulties.

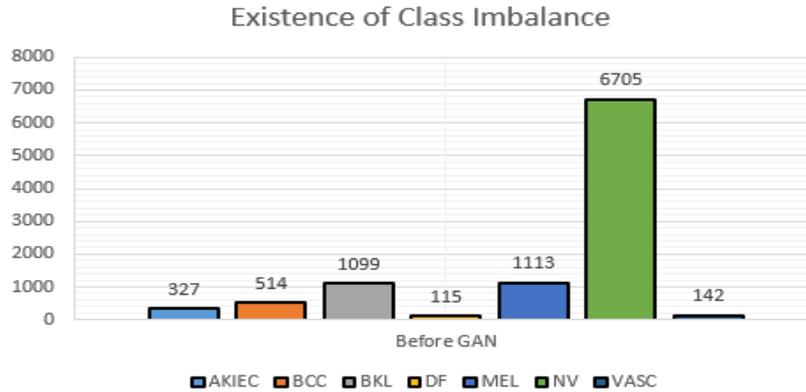

Fig 3. The Existence of Class Imbalance in the Dataset

### 3.2. Data Augmentation

The proposed framework uses GANs to produce synthetic data to overcome the challenges in the dataset. In order to balance the number of images in each class, the real and synthetic images are linked together and further processed by conventional data enhancement techniques such as flipping, rotating, etc. This subsection deals with the data augmentation techniques utilized to minimize class imbalance.

### 3.3. Generative Adversarial Networks

LesionAid proposes the utilization of VITGAN to produce synthetic images from the original skin lesion images. The VITGAN is utilized to be compatible in terms of performance and computational efficiency. The recently obtained synthetic and original images are further labeled and employed to train the neural network models. We use the HAM 10000 dataset images to train the generative adversarial networks (GANs). GANs are certain types of architectures composed of CNNs, and GANs mainly consist of two parts: Discriminator (D) and Generator (G). This subsection explains the architecture of the coupled generator and discriminator model in our VITGAN. The complete architecture of generator and discriminator of the ViTGAN is shown in Fig. 4.

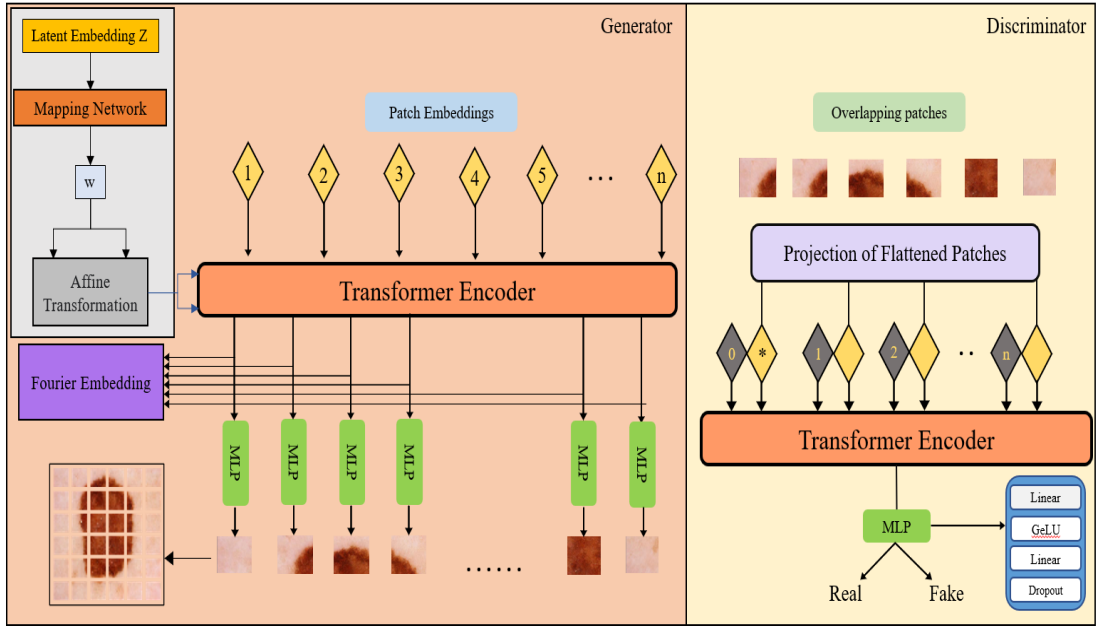

Fig. 4 The Generator and Discriminator Architectures in the Proposed ViTGAN

### 3.4. ViT GAN Discriminator Architecture

In recent times, Lipschitz continuity has become prevalent in generative adversarial networks (GANs). The Lipschitz regularization in GANs enhances the training resilience and sample quality. The conventional executions of Lipschitz continuity in GANs include gradient penalty and spectral normalization. In our work, Lipschitzness of ViT discriminator enforcement is performed with L2 attention proposed in [28]. In contrast to Eq. 1, in the self-attention head, the dot product similarity with Minkowski distance and also concatenates the weights for the projection matrices for queries and keys,

$$SA_Z = softmax\left(\frac{d(W_Q, W_K)}{\sqrt{d_z}}\right) \times W_V \quad (1)$$

$W_Q$, $W_K$, and $W_V$ are the projection matrices for queries, keys, and values. d(a, b) calculates the vectorized L2 distances between a and b points. $d_z$ is the feature dimension of the keys or the scaling factor for each head. This transformation enhanced the stability of ViT when employed for GAN discriminators.

ViT discriminators have a high learning ability, which makes them susceptible to overfitting. The discriminator and generator use the same lesion images, which divide the image into a series of non-overlapping regions according to a specific M × M pattern. Even if not appropriately calibrated, these random grid subdivisions may allow the discriminator to remember local clues and no longer offer significant losses to the generator. We employ a straightforward technique to address this problem by permitting considerable overlapping between image patches. We add pixels to each boundary edge of the patch to increase the actual patch size (M + 2o) × (M + 2o).

### 3.5. ViT GAN Generator Architecture

It is challenging to create a generator using the ViT architecture. Transforming ViT from estimations on a group of class labels to produce pixels over a bounded area is difficult. The proposed ViTGAN Generator consists of two parts:
- Transformer block
- Output Mapping Layer

We employ the noise vector z to modulate the layer norm operation rather than sending it as the input to ViT. The generator's components are initiated to the patch encoders. Because once frames are divided into patches and encoded by a patch encoder, the subsequent vectors are fed into transformer encoders. Stacking N of the transformer encoders through the top of one another, which are utilized to comprehend the background. This perspective in this image-to-image translation is the image. The result of the transformer encoder is transmitted to a residual block. The residual block is composed of two convolutional layers separated by skip connections. In the residual blocks, ReLU and batch normalization are used. Following residual blocks is an upsampling block. In the upsampling block, transpose convolutions are followed by Leaky ReLU and batch normalization.

### 3.6. Positional and Color Data Augmentation

Training and deploying the Vision Transformer models on more image data can result in better accuracy and skillful models. The Image augmentation techniques can generate variations of the images that can enhance the mastery and performance of the ViT frameworks. Image augmentation is a technique to artificially create new training images from existing training images. Image transformations include operations such as shifting, flipping, zooming, etc. Fig. 5 shows the enlargement of an exemplary cut-out face. Table I shows the configuration for data expansion to generate new training image data.

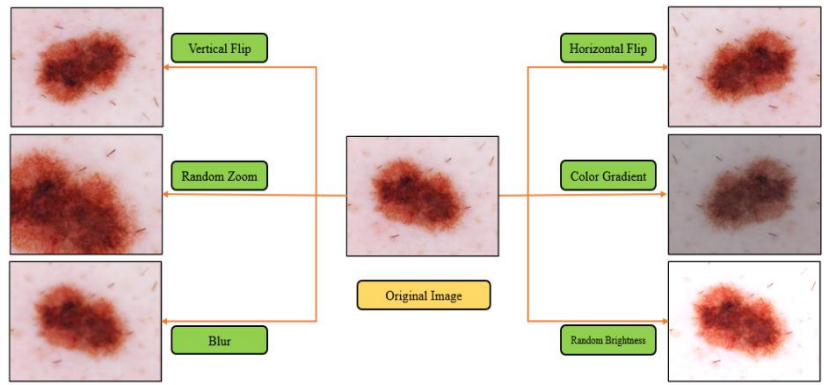

Fig 5. The Conventional Positional-Color Data Augmentation Techniques

**Table I**
**The factors of data augmentation**

| Technique | Hyper-Parameter |
| --- | --- |
| Normalization | Yes |
| Resize | No |
| Random Flip | Horizontal and Vertical |
| Random Rotation | Factor = 0.75 |
| Random Zoom | Factor = 0.05 |
| Random Brightness | Range = [0.01, 0.5] |

### 3.7. Vision Transformers for Image Classification

Vision transformers are a type of deep learning architecture that uses transformer networks for image classification tasks. Like other transformer networks, vision transformers use self-attention mechanisms to process the input image as a sequence of patches rather than as a grid of pixels; this allows the network to capture long-range dependencies in the image and improve its performance on image classification tasks. Vision transformers have been shown to achieve state-of-the-art results on several image classification benchmarks, such as ImageNet. They can perform well on various image classification tasks, including object recognition and scene classification. In addition to their impressive performance, vision transformers have several advantages over other deep learning architectures for image classification. For example, they can be trained using relatively small amounts of data and handle input images of arbitrary size without needing manual resizing or cropping; this makes them a good choice for image classification tasks in real-world applications.

Once face identification and picture augmentation are completed, image classification requires an effectively working machine learning architecture. Vision transformers are used in modern methodological approaches to classify images appropriately. Without using the framework's convolution layers, the ViT framework uses the transformer model with a series of picture speckles. In contrast to conventional transformers, the ViT architecture receives a sequence of linear embeddings of the separated patches out of an input frame that has been shrunk and transformed into N patches.

$$I \in R^{H \times R \times C} \Rightarrow Ip \in R^{N \times (P^2 \times C)} \quad (2)$$
$$N = H \times \frac{W}{P^2} \quad (3)$$

Here I represent the image and H, W, and C stand for the height, the width, and the channels of the original image, (P, P) for the resolution of the individual spots, and N for the total number of spots created as well as for the representation of the spots created using the example image.

Additionally, each patch's matrices of dimensions (P, P) are flattened into matrices with dimensions (1, $P^2$). These compressed patches (E(1, $P^2$) are transmitted via a single feed-forward layer that comprises an embedding matrix (F) to produce a linear patch projection ($P^2$, D). Following projection, the patches changed into patch embeddings (E), which are fixed latent vectors of size D (projection dimension) (1, D), depicting the creation of patch embeddings and providing a conceptual overview of the ViT model (a). To help the classification head, a learnable [class] embedding is prehended to linear patch projections. The order of succession needs to be implemented naturally since the data is supplied in a single instant. The concatenated matrix consisting of the patch embeddings (E) and the learnable class embeddings ($E_{pos}$) is added to the concatenated matrix to solve this problem ($I_{class}$). The equations supply the extracted patches that were extracted, as a result, embedded sequentially with the token z0.

$$z_0 = [I_{class}; x_1E; x_2E; \ldots\ldots x_nE + E_{pos} \qquad (4)$$

$$E \in R^{(P^2C) \times D}, E_{pos} \in R^{(N+1) \times D} \qquad (5)$$

The input for the transformer encoder, which is made up of L identical layers of the Multi Headed Self-Attention (MSA) blocks and Multi-Layer Perceptron (MLP) blocks as illustrated, are these embedding sequences combined. According to mathematical equations, both transformer encoder subcomponents function with residual skip connections after the normalization layer (LN).

$$z'_l = MSA(LN(z_{l-1}) + z_{l-1} \qquad (6)$$

$$z_l = MLP(LN(z'_l) + z'_l \qquad (7)$$

The MSA block's associated operations and the input vectors are stacked and multiplied by a starting set of weights Wq, Wk, and Wv in the attention block to create three different matrices containing queries Q, keys K, and values V. The attention matrix is created by multiplying the matrix Q by the transpose of the matrix K, resulting in a dot product of each query q from Q and each key k from K. Although it includes the dimension of the key dk as a scaling factor, the scaling dot product used in the Self Attention (SA) block is identical to the standard dot product. The softmax is fed with the dot product results to determine the attention weight. The scalable dot product attention for each h head is calculated using the MSA block in the transformer encoder. Concatenated results from each attention head are then sent to a feed-forward layer with learnable weights $W^0$ as represented in the equations.

$$SA = Softmax(QK^T \sqrt{d_k}) \times V = W_{attention} \times V \qquad (8)$$

$$MSA = concat(SA_1, SA_2, SA_3, \ldots\ldots, SA_h) \times W^0 \qquad (9)$$

$$w^0 \in R^{hd_k \times D} \qquad (10)$$

The feed-forward dense layers with GeLU highly nonlinear make up the MLP block. The top element in the sequence ZL is sent to an external lead classifier at the last layer of encoding to estimate the class labels. The architecture of the proposed ViT model along with the transformer encoder and multi-attention block is shown in the fig 6.

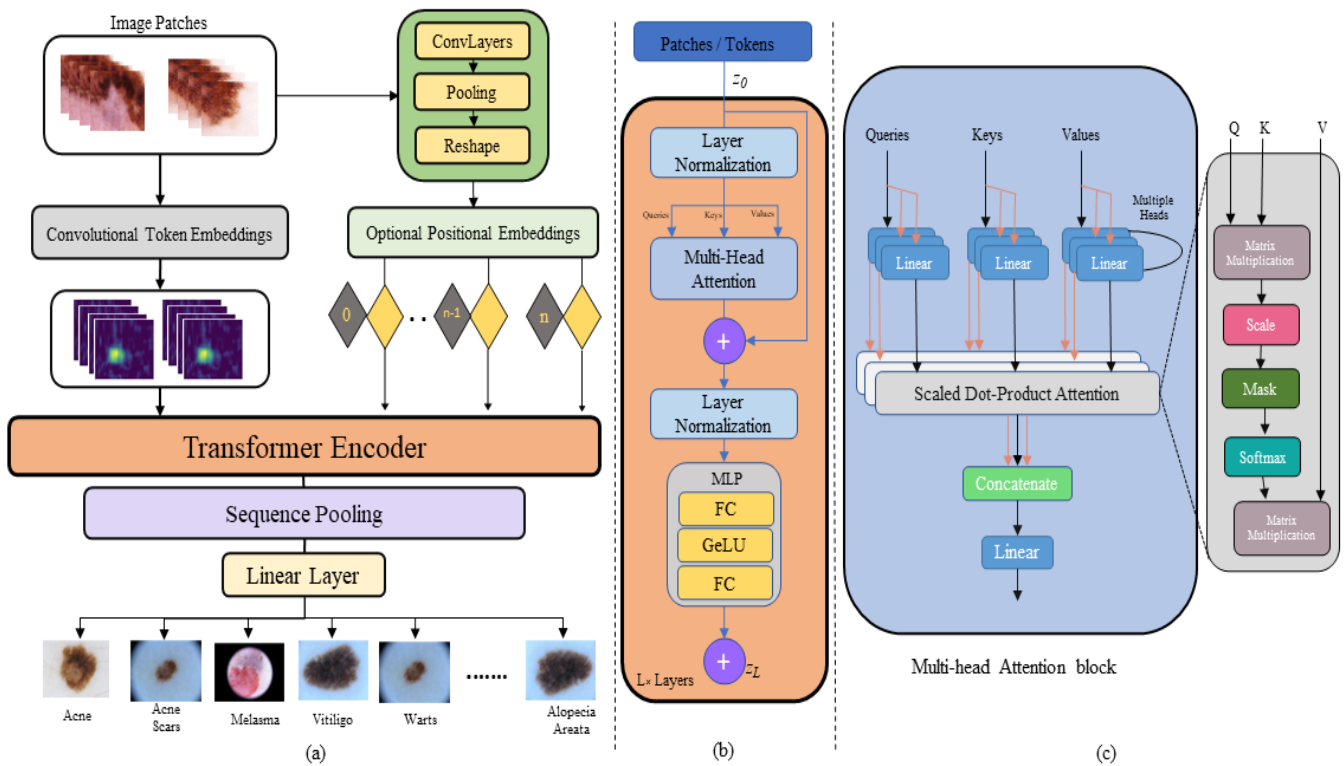

Fig. 6 (a) Conceptual Overview of ViT Model (b) Transformer Encoder (c) Multihead Attention Block

## 3.8. Explainable AI

Explainable AI describes approaches and methodologies in AI that allow human researchers to comprehend the outcomes of the solution. In the early stages of AI adoption, it was acceptable to only partially comprehend the model's predictions as long as the outputs were accurate. The explanation of their operation was not the top priority. Now, the emphasis is on developing human-interpretable models. It differs from the "black box" notion in machine learning, in which not even the AI's creators can explain why it made a particular decision. XAI is a realization of the social right to clarify. AI can optimize choices, which positively and negatively affect businesses. Similar to hiring organizational decision-makers, it is crucial to comprehend how AI makes choices. Numerous organizations desire to utilize AI but also are hesitant to allow the framework or AI to have more effect at a time because they still need to trust the framework. Explainability facilitates this because it reveals how models make decisions.

The Explainable AI (XAI) program aims to develop a collection of machine learning methods that:
  i. Generate more explainable frameworks while retaining a high level of learning performance (prediction accuracy); and
  ii. Facilitate humans to comprehend the evolving development of advanced AI collaborators, trust them adequately, and manage them effectively.

## 3.9. Deployment of Web Application

The skin care website consists of the following folder structure, which you can see in its entirety below. Explanation of the folder structure that is included in the Python Flask framework, as seen in Figure 6:
- Templates, also known as views, are responsible for handling the display logic. This portion of a web application is often an HTML template file, and the controller is the one who determines its contents. The user is the primary focus of the view block functionality.
- A direct connection to the model section is unavailable via the section views.
- The model often deals directly with databases to change data (insert, update, delete, and search). It handles validation from the controller portions; however, the model cannot interact directly with the display section.
- The controller is a component responsible for regulating the interaction between the model and views parts. The controller's tasks include receiving requests and data from the user and deciding what the program should do with this information.
- You can use Static to store things like CSS and JavaScript and photos.

## 4. Experimental Analysis

### 4.1. Computer Specification of the Proposed System

This section deals with the hardware and software computational specifications of our LesionAid framework. Pytorch, Tensorflow 2.0, Keras, and OpenCV libraries were used to build the framework in Python 3.9. The VITGAN and ViT models are trained using a high-performance graphics processing unit but can also be trained with low-end computational units. The specifications and minimum requirements for training and testing the framework are depicted in Table II.

**TABLE II**
**The Computational Specifications and Requirements**

| Specifications | Our System's Configuration |
|---|---|
| Operating System | Windows 11 |
| CPU | Intel i9 11th Gen |
| RAM | 31.6 Usable |
| GPU | Nvidia RTX 3070 Ti |
| Frameworks | Pytorch, Tensorflow, OpenCV |

### 4.2. Dataset Preparation and Utilization

This experiment utilizes a skin lesion benchmark dataset, HAMS10000 [10], one of the most extensive datasets among available public datasets, for training the predictive models for skin lesion classification. The training HAMS10000 dataset includes 10000 images of 7 different skin lesion diseases. The resolution of the frames in this dataset is $600 \times 450$, which is relatively high compared to other skin disease detection datasets. The skin lesions in the dataset have significant transformations in scale, position and perceptions. Thus, this benchmark dataset is appropriate to demonstrate efficiency and performance in practical scenarios. Though there exists a class imbalance in the data, i.e., the NV class has 6705 lesion images, and MEL class has only 142 lesion images. This existence of class imbalance leads to over-classify the majority group due to its increased prior probability. To compensate for this problem, the LesionAid uses data augmentation techniques such as GANs and positional and color-based augmentation. The difference between our customized dataset and the HAMS10000 dataset is given below in Fig.7.

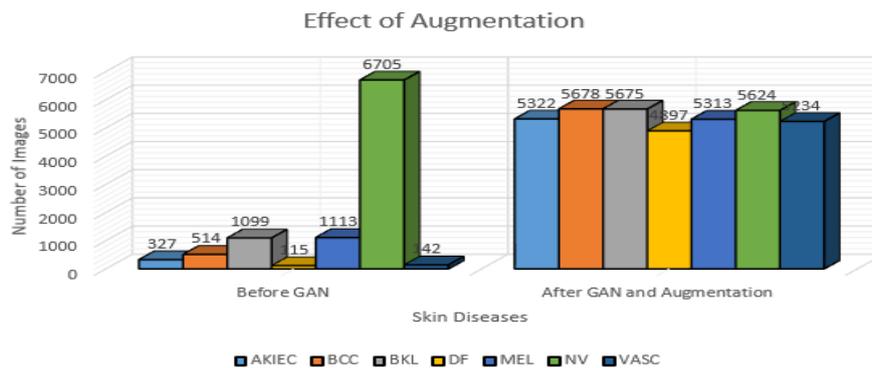

Fig. 7 The Effect of Image Augmentation in the Proposed ViTGAN

### 4. 3. Performance of Data Augmentation

The evaluation of the quality level of synthetic data still needs to be answered. Diversity and both broad- and fine-grained nuances are frequently underrated by the available measurements. Most commonly used measures, including FID and Inception Score, rely on ImageNet-trained networks to extract attributes, making them suspect for cases where classes are incredibly diverse, like those in the medical field. We solve that problem by doing a thorough analysis that includes both conventional GAN metrics on trained classifier models and visualization methods.

**4.3.1 Performance of ViT GAN**

We considered both the amount of training duration and the FID score to choose the ideal training checkpoint for the GAN model, favoring later checkpoints (more extended training) for comparable FID. Frechet inception distance (FID) is a metric used to evaluate the performance of generative models. It measures the distance between the distribution of authentic images and the distribution of generated images from a generative model. FID is often used as a benchmark for comparing the performance of different generative models, as it provides a single scalar value that summarizes the quality of the generated images. The precise training durations, and associated FIDs, of different GAN architectures are shown in Fig. 8. Furthermore, the comparison of the details of the images generated by the proposed framework and existing architectures is shown in Fig.9.

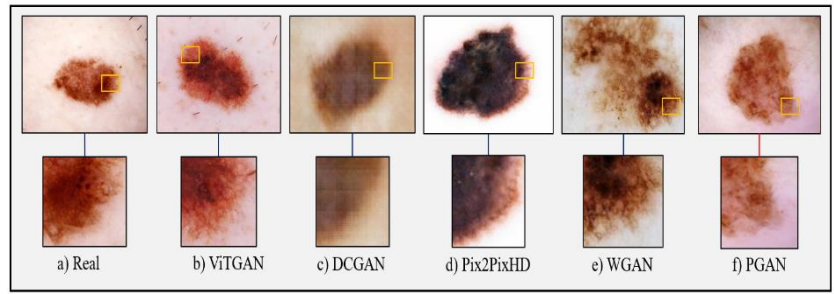

Fig. 8 The Comparative Analysis of Artifacts with Existing Architectures

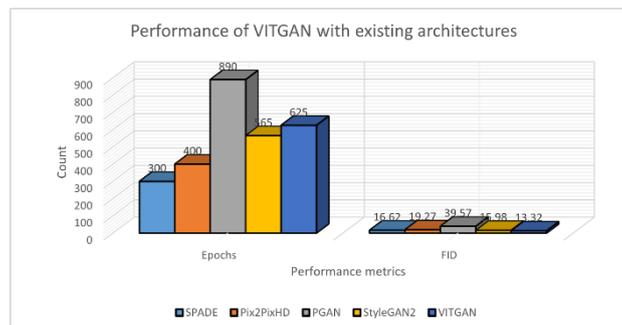

Fig. 9 Comparison of Performance of ViT GAN with Existing Architectures

**4.4. Performance of ViT for Image Classification**

The ViT model is trained to classify multi-skin diseases in this experiment. There are nine possible classes: Warts, vitiligo, acne scars, acne, oily, acanthosis nigricans, dry, melasma, and alopecia-areata. The ViT model is trained to classify multi-skin diseases in this experiment. In order to evaluate our ViT system, the standard standards for assessing classification models have been used. Figure 10 represents the learning curves of accuracy and loss for the training and validation of the models. These learning plots are evidence of an effective learning algorithm since the validation and training curves both retain a point of stability with a minimum difference between them. In order to get the best possible results, the training of the effective ViT model was designed to integrate three separate but interrelated tasks simultaneously: 1) the calculation of output, 2) the correction of mistakes, and 3) the fine-tuning of the hyper-parameters. Following several rounds aimed at fine-tuning the hyper-parameters, the highest possible training and validation accuracy was found to be 99.2 percent and 97.4 percent, respectively.

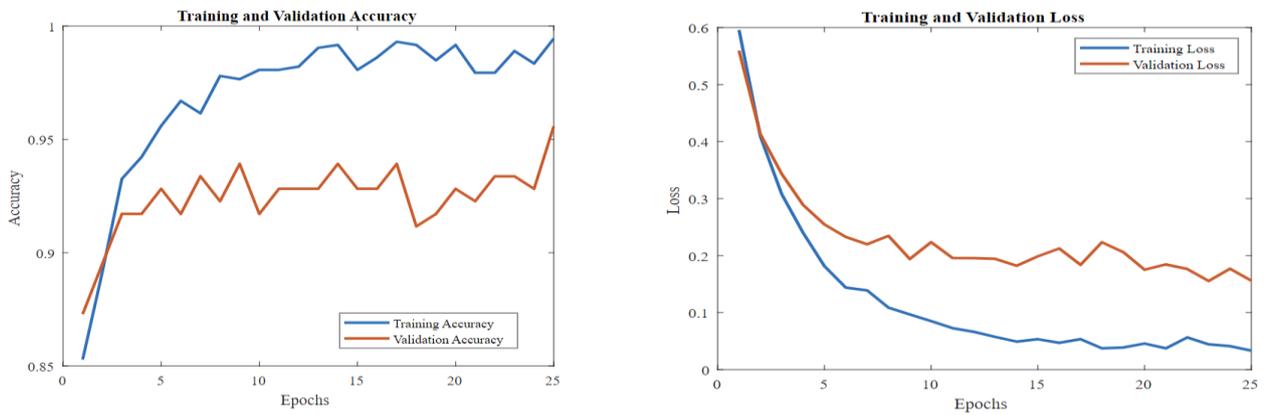

Fig 10. Learning Curves for Training and Validation Accuracy for 25 Epochs

**4.5. Explainable AI**

Explainable AI (XAI) includes methods for explaining the reasoning behind a particular decision or action or for providing an understanding of how the AI system works. Gradient-weighted Class Activation Mapping (GradCAM) is a technique for visualizing the regions of an input image that are most important for a convolutional neural network (CNN) to make a prediction. It is a gradient-based attribution method, which uses the gradients of the output of the CNN concerning the input image to identify the critical regions. GradCAM works by first using CNN to predict the input image. The gradients of the output of the CNN concerning the feature maps in the last convolutional layer are then computed. These gradients are used to weight the feature maps, and the weighted feature maps are then averaged to produce a heatmap highlighting the regions of the input image most important for the prediction. GradCAM can be used to understand better how a CNN is making predictions and can help to identify errors or biases in the model. It can also help debug and improve the performance of a CNN. Additionally, GradCAM can generate attention maps, which can be helpful for tasks such as image captioning or visual question answering. The GradCAM attention maps for the lesion images is shown in the fig 11.

**4.6. Web Application**

Dermatologists may determine the type of lesion using this application, which serves as a diagnostic tool. The user (the dermatologist) will first locate the lesion on his body. Then he or she will employ the android camera/ any camera to record the lesion, with the data saved to the edge device by our website. After that, the user launches the Lesion application and selects the readily apparent symptoms. This stage is referred to as the symptom analysis phase. In the next stage, the user will acquire the class of the lesion after uploading a photograph of the lesion. Figures 12, 13, 14 depict the interface of the proposed website. To utilize the proposed web application, the user needs to follow a few regulations to obtain efficient and accurate results for their input images.

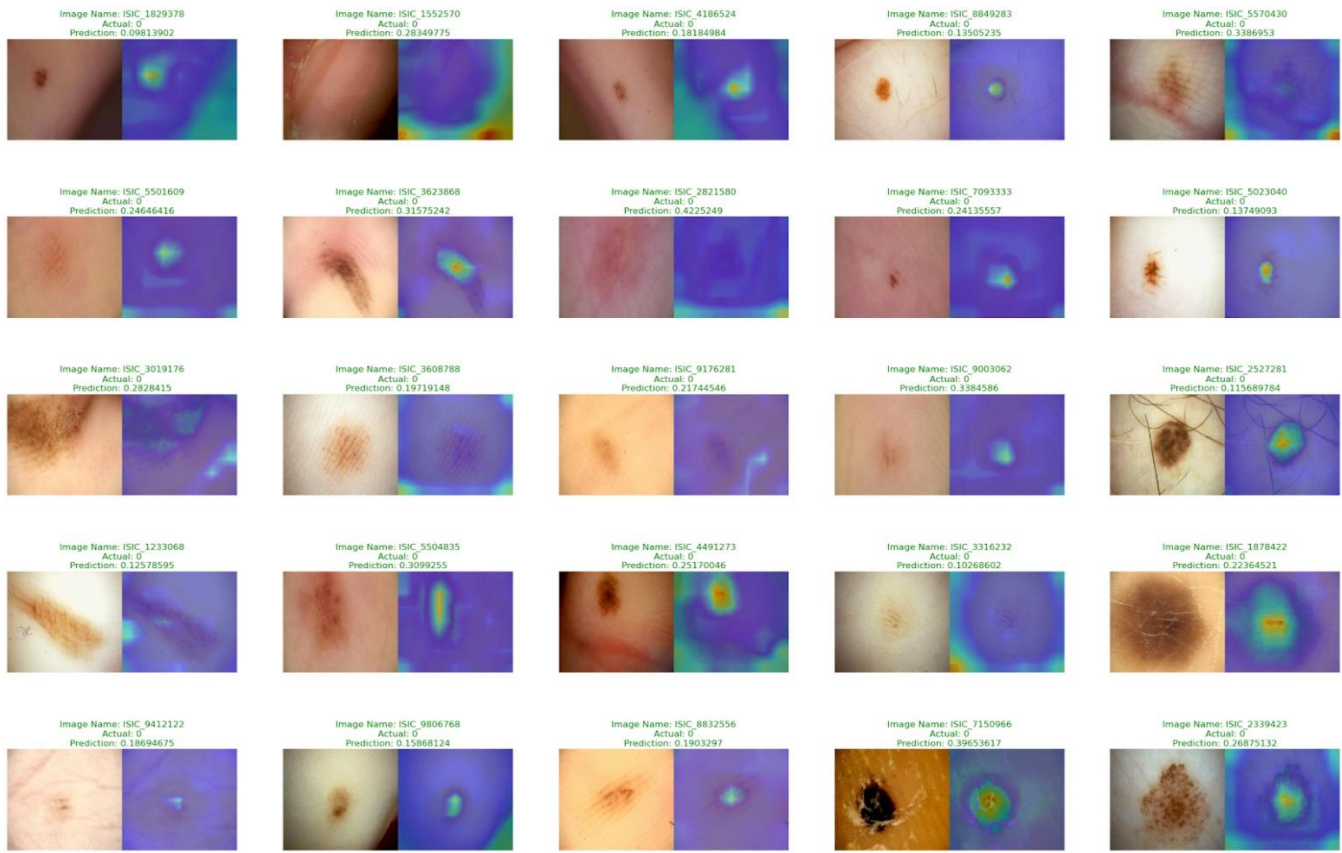

Fig 11. Attention Maps Using GradCAM Technique

**Table III**

**The input image requirements**

| Specifications | Our System's Configuration |
|---|---|
| Image Size | Preferably 615:512 pixels |
| Aspect Ratio | 4:3 |
| File Size | 20–200 Kb |
| Brightness | 200 Nits |
| Dots Per Inch (DPI) | 80 dpi |

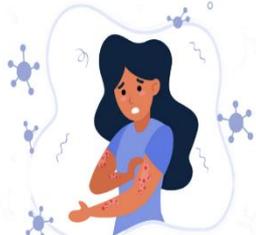

Fig 12. Home Page of the Website

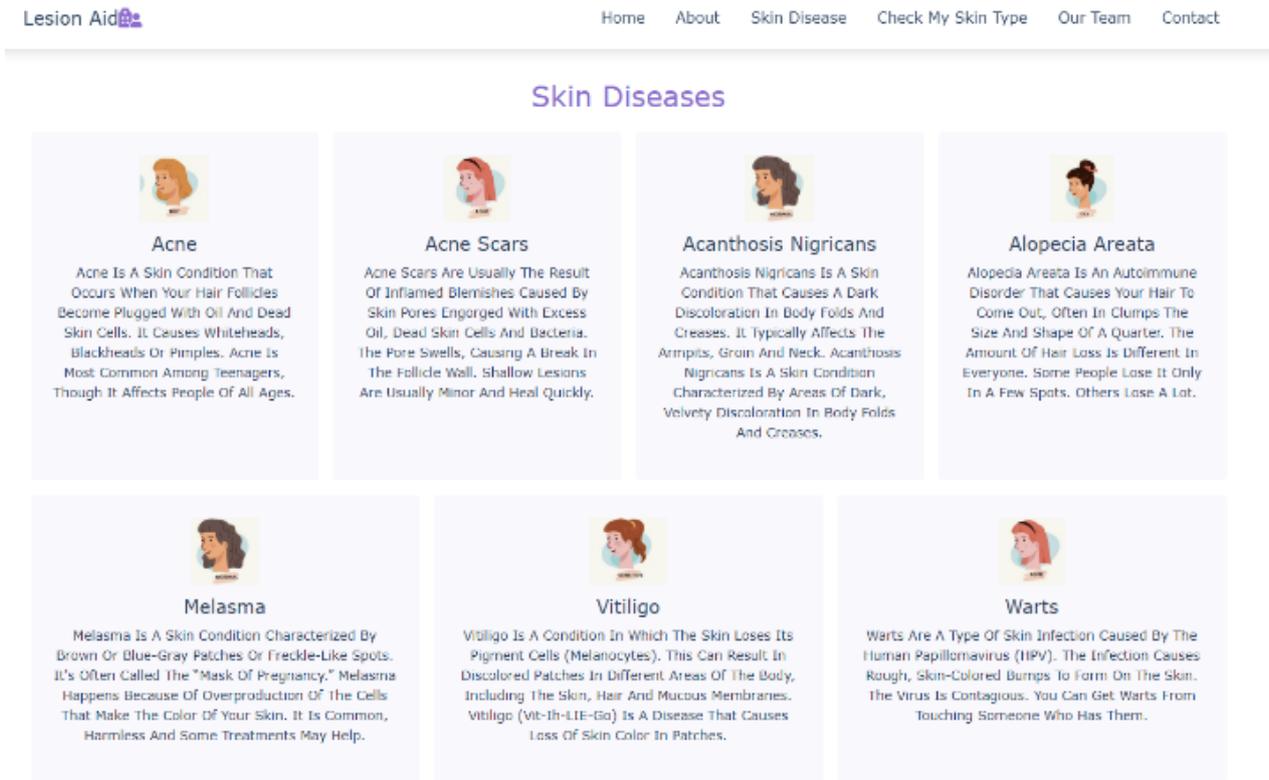

Fig 13. Seven different classes that can be classified using the proposed model

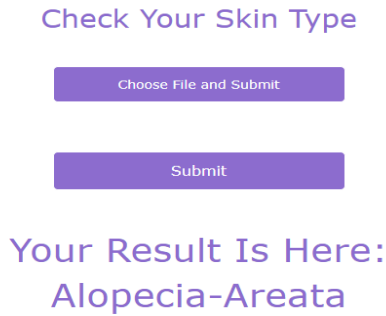

Fig 14. Giving the user input and predictions by the Framework

## 5. Conclusion

We have developed Vision Transformers for image/lesion classification in this work. Concurrently, GAN is based on a generator and discriminator model that generates synthetic images to eliminate the class imbalance issue. Thus, we have developed an end-to-end framework for data augmentation using ViT GAN, image classification using ViT, explainable AI for improved model comprehension, and a web application for end users to predict their lesion type. More experiments have been conducted to evaluate the performance of GAN and determine how to make it more stable to enhance its performance. We have used the FID score as a training benchmark for the GAN model. For 625 epochs, the ViT obtained an FID score of 13.36. We have used positional and color data augmentation because it is equally important to test with different types of data with many variations in terms of illumination, color, and texture, among others. Then, a ViT model is constructed to classify the lesions. In this study, seven lesion types were considered, and the ViT model achieved a training and validation accuracy of 99.2% and 97.4%, respectively. Explainable AI has been implemented to comprehend the reasoning behind a specific decision or action or to provide an understanding of how the AI system operates. GradCAM is a technique used to visualize the input region to predict the lesion with the ViT model.

# REFERENCES

[1] K. Roy, S. S. Chaudhuri, S. Ghosh, S. K. Dutta, P. Chakraborty, and R. Sarkar, "Skin Disease detection based on different Segmentation Techniques," 2019 International Conference on Opto-Electronics and Applied Optics (Optronix), 2019, pp. 1-5, DOI: 10.1109/OPTRONIX.2019.8862403.

[2] A. Ajith, V. Goel, P. Vazirani, and M. M. Roja, "Digital dermatology: Skin disease detection model using image processing," 2017 International Conference on Intelligent Computing and Control Systems (ICICCS), 2017, pp. 168-173, DOI: 10.1109/ICCONS.2017.8250703.

[3] A. Haddad and S. A. Hameed, "Image Analysis Model for Skin Disease Detection: Framework," 2018 7th International Conference on Computer and Communication Engineering (ICCCE), 2018, pp. 1-4, DOI: 10.1109/ICCCE.2018.8539270.

[4] S. Kolkur and D. R. Kalbande, "Survey of texture-based feature extraction for skin disease detection," 2016 International Conference on ICT in Business Industry & Government (ICTBIG), 2016, pp. 1-6, DOI: 10.1109/ICTBIG.2016.7892649.

[5] Quan Gan, and Tao Ji, Skin Disease Recognition Method Based on Image Color and Texture Features, Computational and Mathematical Methods in Medicine / 2018 / Article, Volume 2018.

[6] M. N. Alam, T. T. K. Munia, K. Tavakolian, F. Vasefi, N. MacKinnon and R. Fazel-Rezai, "Automatic detection and severity measurement of eczema using image processing," 2016 38th Annual International Conference of the IEEE Engineering in Medicine and Biology Society (EMBC), 2016, pp. 1365-1368, DOI: 10.1109/EMBC.2016.7590961.

[7] A. Romero Lopez, X. Giro-i-Nieto, J. Burdick and O. Marques, "Skin lesion classification from dermoscopic images using deep learning techniques," 2017 13th IASTED International Conference on Biomedical Engineering (BioMed), 2017, pp. 49-54, DOI: 10.2316/P.2017.852-053.

[8] A. Mahbod, G. Schaefer, C. Wang, R. Ecker, and I. Ellinge, "Skin lesion classification using hybrid deep neural networks," Proc. ICASSP 2019 - 2019 IEEE International Conference on Acoustics, Speech and Signal Processing (ICASSP), pp. 1229–1233, 2019.

[9] Parvathaneni Naga Srinivasu, Jalluri Gnana SivaSai, Muhammad Fazal Ijaz, Akash Kumar Bhoi, Wonjoon Kim and James Jin Kang Classification of Skin Disease Using Deep Learning Neural Networks with MobileNet V2 and LSTM, Sensors 2021, 21(8), 2852.

[10] Z. Qin, Z. Liu, P. Zhu, and Y. Xue, ''A GAN-based image synthesis method for skin lesion classification,'' Comput. Methods Programs Biomed., vol. 195, Oct. 2020, Art. no. 105568.

[11] Maleika Heenaye-Mamode Khan, Nuzhah Gooda Sahib-Kaudeer, Motean Dayalen, Faadil Mahomedaly, Ganesh R. Sinha, Kapil Kumar Nagwanshi, Amelia Taylor, "Multi-Class Skin Problem Classification Using Deep Generative Adversarial Network (DGAN)," *Computational Intelligence and Neuroscience*, vol. 2022, Article ID 1797471, 13 pages, 2022.

[12] P. R. Medi, P. Nemani, V. R. Pitta, V. Udutalapally, D. Das and S. P. Mohanty, "SkinAid: A GAN-based Automatic Skin Lesion Monitoring Method for IoMT Frameworks," 2021 19th OITS International Conference on Information Technology (OCIT), 2021, pp. 200-205, doi: 10.1109/OCIT53463.2021.00048.

[13] H. Rashid, M. A. Tanveer and H. Aqeel Khan, "Skin Lesion Classification Using GAN based Data Augmentation," 2019 41st Annual International Conference of the IEEE Engineering in Medicine and Biology Society (EMBC), 2019, pp. 916-919, doi: 10.1109/EMBC.2019.8857905.

[14] Elngar, Ahmed & Kumar, Rishabh & Hayat, Amber & Churi, Prathamesh. (2021). Intelligent System for Skin Disease Prediction using Machine Learning Intelligent System for Skin Disease Prediction using Machine Learning. Journal of Physics Conference Series. 1998. 12037. 10.1088/1742-6596/1998/1/012037.